\documentclass{article}

\newcommand{\FIG}[1]{\textbf{\autoref{fig:#1}}}
\newcommand{\TABLE}[1]{\autoref{tab:#1}}

\newcommand{\toprule}{\hline}
\newcommand{\midrule}{\hline}
\newcommand{\bottomrule}{\hline}

\usepackage{arxiv}

\usepackage[utf8]{inputenc} 
\usepackage[T1]{fontenc}    
\usepackage{hyperref}       
\usepackage{url}            
\usepackage{booktabs}       
\usepackage{amsfonts}       
\usepackage{nicefrac}       
\usepackage{microtype}      
\usepackage{lipsum}
\usepackage{graphicx}
\usepackage{rotating}
\graphicspath{{figs/}}
\title{Supervised dimensionality reduction by a Linear Discriminant Analysis on pre-trained CNN features}

\author{
  Francisco J. H.~Heras \\
 Champalimaud Research\\
  Champalimaud Center for the Unknown\\
  Lisbon, Portugal \\
  \texttt{francisco.heras@neuro.fchampalimaud.org } \\
   \And
 Gonzalo G.~ de Polavieja \\
   Champalimaud Research\\
  Champalimaud Center for the Unknown\\
  Lisbon, Portugal \\
  \texttt{gonzalo.polavieja@neuro.fchampalimaud.org} \\
 \\
}

\begin{document}
\maketitle

\begin{abstract}
We explore the application of linear discriminant analysis (LDA) to the features obtained in different layers of pretrained deep convolutional neural networks (CNNs). The advantage of LDA compared to other techniques in dimensionality reduction is that it reduces dimensions while preserving the global structure of data, so distances in the low-dimensional structure found are meaningful. The LDA applied to the CNN features finds that the centroids of classes corresponding to the similar data lay closer than classes corresponding to different data. We applied the method to a modification of the MNIST dataset with ten additional classes, each new class with half of the images from one of the standard ten classes. The method finds the new classes close to the corresponding standard classes we took the data form. We also applied the method to a dataset of images of butterflies to find that related subspecies are found to be close. For both datasets, we find a performance similar to state-of-the-art methods.
\end{abstract}


\section{Introduction}

A common approach when confronting high-dimensional data is to find a low-dimensional structure formed by the points of interest. For example, portraits are a tiny subset of all grayscale images, and we can use techniques (e.g. VAE, UMAP) to find a low-dimensional manifold that passes next to them. This produces an embedding: to identify each portrait image, instead of pixel intensities (many numbers), we can specify its coordinates in the lower-dimensional manifold. In addition, we might want the embedding to lose some information about pose, light, etc, and to map all images from the same person into the same point. Once we have this embedding, we can verify face identities by thresholding distances, and recognition of new individuals by clustering \cite{schroff2015facenet}.  Experimentally, these embeddings map images of people that look alike to neighbour points in the embedding. We could try to assess similarity between people by measuring distances in the embedding. However, there is no guarantee that the global structure of the space is preserved by the embedding.

An alternative is linear discriminant analysis (LDA), a technique that reduces dimensions while preserving the global structure of data \cite{hastie2009elements}. In this transformed space, each dimension is scaled so the distribution of data in each class is as close as possible to an standard normal distribution. Distances between objects in this space are meaningful, since the space is homogeneously deformed, and length along each dimension is expressed in multiples of intra-class variation. 

LDA can use pixel intensity information, but we can improve the generated embedding by providing image features instead. Deep convolutional neural networks (CNNs) are the state-of-the-art neural networks to perform image classification. The hypothesis is that once trained to perform classification in large image datasets, CNN learn to obtain useful features from natural images. For instance, it is a common practice to use the first layers of a pretrained CNN, and train a classifier on top (e.g. \cite{sharif2014cnn, el2016face, lopes2017pre}). In addition, CNNs were found to discover similar features than biological visual systems \cite{verma2014using}, and have been considered as models of animal visual systems \cite{lindsay2020convolutional}.

Here we present a study that explores the embeddings created by applying LDA to the features obtained in different layers of pretrained deep CNNs. In Section 2 we give detailed information about the networks and methods used. In Section 3.1 and Section 3.2 we apply CNN-LDA to two different datasets. In Section 4 we conclude that CNN-LDA is potentially useful for supervised dimensionality reduction.

\section{Methods}
\label{sec:methods}

We use a pre-trained CNN to obtain features of an image in the following way. Given a pre-trained CNN, truncated at an intermediate layer, and an image, we perform a forward pass of the image. The flattened (one-dimensional) result is the vector of convolutional features of the image that we use for further analysis. We perform this forward pass for all images in the training set. 

The number of features of each image can be more than tens of thousands. We did not consider any layer with more than $15,000$ features. In a variant of this method, we use principal components analysis (PCA) as an intermediate step to reduce the number of features of each image, before performing LDA. When performing PCA as a preprocessing step, we considered layers up to $25,000$ features.

We then use linear discriminant analysis (LDA) to perform dimensionality reduction and classification. To perform LDA, we use algorithms available in scikit-learn \cite{scikit-learn}. The dimension of the space is the number of classes in the dataset minus one. The mapping of the images into that space is the embedding we study. We validate the quality of the embedding by using a held out test dataset.


\subsection{Networks used}

We selected 6 architectures with high accuracy to number of parameters ratio \cite{bianco2018benchmark}: DenseNet \cite{iandola2014densenet}, GoogLeNet \cite{szegedy2015going}, MnasNet \cite{tan2019mnasnet}, MobileNetv2 \cite{sandler2018mobilenetv2}, ShuffleNetv2 \cite{ma2018shufflenet} and SqueezeNet1.1 \cite{iandola2016squeezenet}. We used weights obtained by training on ImageNet. We downloaded them from torchvision, part of the pytorch ecosystem \cite{pasze2019pytorch}. 

To obtain convolutional features, we have the option to truncate each CNN at different layers. For each network we enumerated some of these possible truncations and used the features generated at these points. We follow the upper hierarchy of the pytorch definition, which often correspond to points in the network where the number of features is small. 

\subsection{Linear discriminant analysis}

The assumption of linear discriminant analysis is that data is generated from a Gaussian Mixture, each class modelled as a multivariate normal, with all sharing the same covariance matrix (section 4.3 in \cite{hastie2009elements}). During learning, the class probabilities, $\pi_i$, the class means, $\mu_i$ and the common covariance matrix, $\Sigma$, are obtained from their sample estimates. Bayes inference of the most probable class for a new data point $x \in \mathbb{R}^p$ is equivalent to finding the class $i \in 1, ..., K$ that minimises:
\begin{equation}
    \delta_i(x) = \frac1{2}(x - \mu_i)^{T} \Sigma^{-1}(x - \mu_i) - \log \pi_{i}
\end{equation}

$\Sigma$ is a real symmetric matrix, so it has an eigendecomposition
\begin{equation}
    \Sigma=U D U^{T}
\end{equation}
where $U \in \mathbb{R}^{p \times p}$ with orthogonal columns and rows and $D$ is a diagonal matrix with non-negative elements. 

Using the decomposition
\begin{equation}
    \delta_i(x) = \left\| D^{-1/2}U^{T} x - D^{-1/2}U^{T} \mu_i \right\|^2_2 - \log \pi_{i} = \left\| \hat{x} - \hat{\mu}_i \right\|^2_2 - \log \pi_{i},
\end{equation}
where we denote with a hat the variables transformed by $D^{-1/2}U^{T}$. Let us consider the $K-1$ dimensional affine subspace $M \subseteq \mathbb{R}^p$, and denote by $P_M$ the projection onto $M$. Each point $\hat{x}$ has an unique descomposition $\hat{x} = P_M \hat{x} + P_{M^{\perp}\hat{x}}$, which produces:
\begin{equation}
    \delta_i(x) = \left\| \hat{x} - \hat{\mu}_i \right\|^2_2 - \log \pi_{i} = \left\| (P_M \hat{x} - \hat{\mu}_i) + P_{M^{\perp}}\hat{x} \right\|^2_2 - \log \pi_{i} = \left\| P_M \hat{x} - \hat{\mu}_i \right\|^2_2 + \left\|P_{M^{\perp}}\hat{x} \right\|^2_2 - \log \pi_{i}.
\end{equation}

Since $\left\|P_{M^{\perp}}\hat{x} \right\|^2_2$ does not depend on the class index $i$, the LDA classification can be performed by finding the class $i$ with minimum 
\begin{equation}
    \delta'_i(x) = \left\| P_M \hat{x} - \hat{\mu}_i \right\|^2_2 - \log \pi_{i}.
\end{equation}

The previous equation says that, once the space is scaled so the intra-class variation is a standard normal (sphering) and projected in the affine subspace defined by the sphered class centroids, classification (except for a correction $\log \pi_{j}$ due to class imbalance) consists in finding the closest class centroid. We use the composition, understood as a map from $\mathbb{R}^{p}$ to $\mathbb{R}^{K-1}$, to define our embedding. 


\subsection{Model selection}

In this limited study, for compatibility with the butterfly dataset, which has few datapoints, we used the test set for model selection. We used the test dataset to measure overfitting for different models (combination of CNN and layer). In the ideal situation, the centroid of the training set embedding and the centroid of the test set embedding would fall in the same point. Any systematic deviation would show overfitting in the embedding learning. We would consider that an embedding overfits less if the average of the distances between the centroids obtained from the training set and the centroids obtained from the test set are small. 
\begin{equation}
    d = \left\langle \sum_j \left( x_{i, j} - y_{i, j} \right)^2  \right\rangle_{i=1.. K},
\end{equation}
where $x_{i, j}$ is the dimension $j$ of the centroid of class $i$ in the embedding (calculated when training LDA). $y_{i, j}$ is the dimension $j$ of the centroid of the embeddings of class $i$ test images.  

Alternatively, we tested a corrected version of this model selection method. In this corrected version, when calculating the distance, we normalise each dimension by the standard deviation of centroids along that dimension:
\begin{equation}
    d_{\textnormal{corrected}} = \left\langle \sum_j \left( \frac{x_{i, j} - y_{i, j}}{SD(x_{i, j}, i=1.. K)} \right)^2 \right\rangle_{i=1.. N} .
\end{equation}

\subsection{Confusion metric}

We quantify separation between different classes of pairs by the sum of false positive rate and false negative rate of the threshold classifier optimally separating the two distributions. If the two distributions are completely overlapping, any threshold will misclassify a fraction $x$ of the first distribution and 1-$x$ of the second distribution (confusion 1). If the two distributions are non-overlapping, a threshold between them will classify all samples correctly (confusion 0).


\section{Results}
\label{sec:results}

We tested our approach in two datasets and compare the results against supervised UMAP and ButterflyNet.

\subsection{MNIST-20 dataset}

We tested the method in a variant of MNIST in which we use two different labels for each digit. The training and test datasets are the original MNIST training and test datasets, but with half of its labels being the original label plus $10$. This label duplication is performed at random, independently in the training and test sets. The result of the manipulation is that an image corresponding to a label in the original MNIST dataset, sat label $0$, will now either be labelled by a $0$ or by a $10$ in the new dataset. We will call the dataset thus generated MNIST-20.

\begin{figure}
  \centering
  \includegraphics[width=\textwidth]{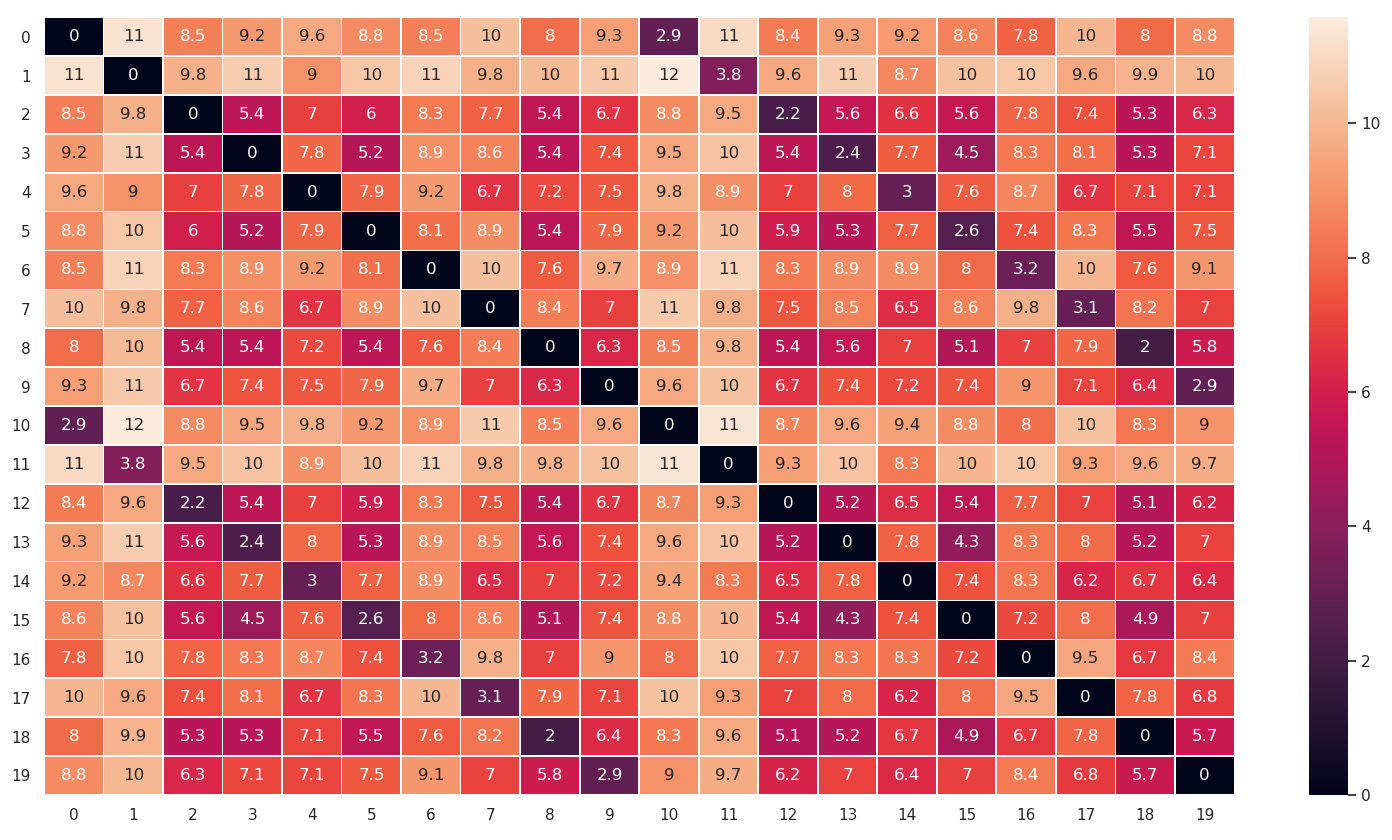}
  \caption{Distances between centroids in MNIST-20. Example result from a single run of CNN-LDA (MobileNet, layer 16). Value at square $(m, n)$ is the distance in the embedding between the centroid of images labelled $m$ and the centroid of images labelled $n$. Note that distances between centroids labelled as $n$ and $n+10$ are small for $n = 0, 1 ... 9$, as expected because they are actually images of the same digit $n$.}
  \label{fig:mnist20best}
\end{figure}

\begin{sidewaysfigure}
  \centering
  \includegraphics[width=\textwidth]{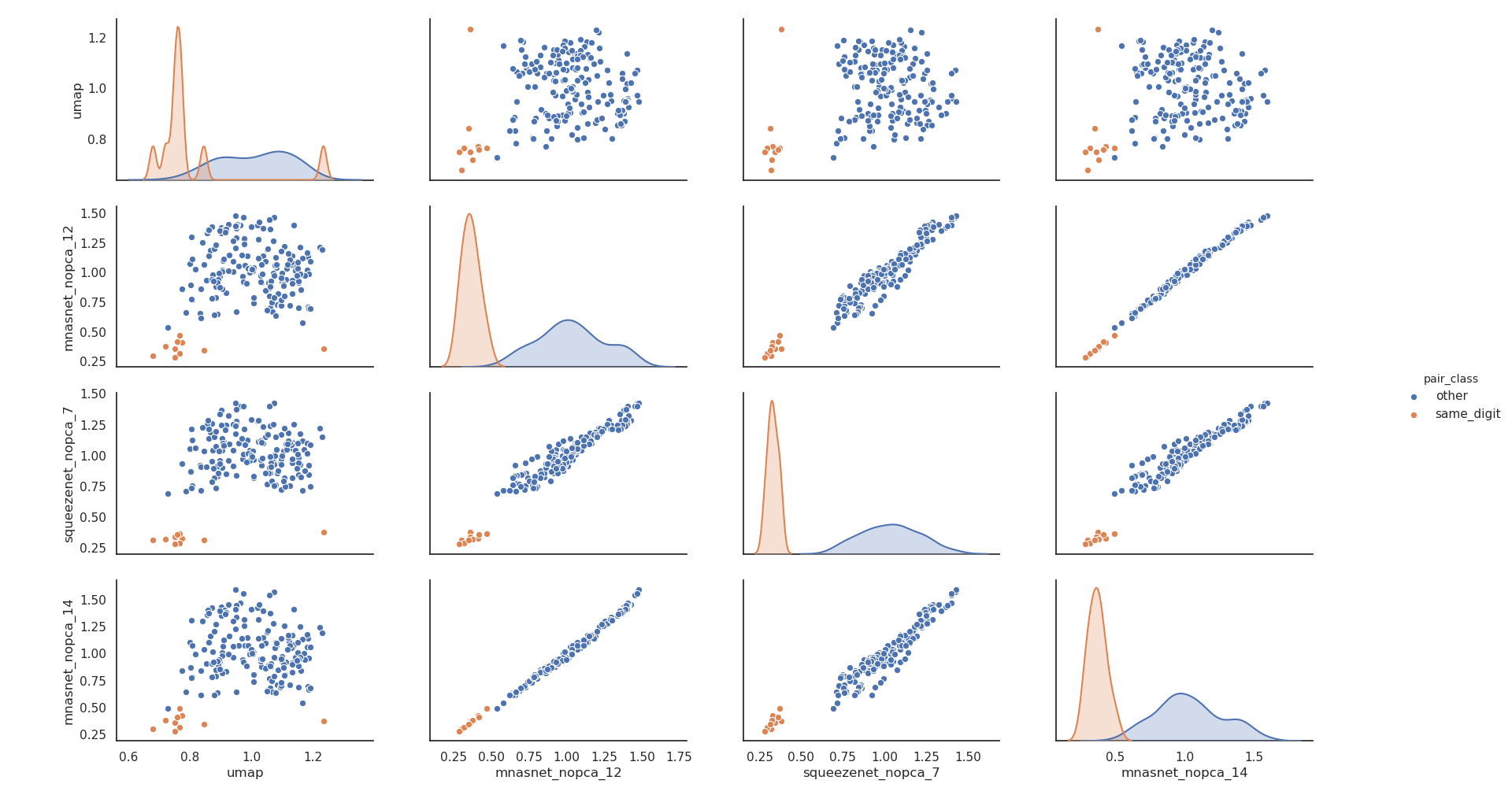}
  \caption{Distances between centroids in MNIST-20. (off-diagonal) Each point represents distances between two centroids, calculated from different embeddings in x and y axis. If the centroids correspond to the class same\_digit (the images represent the same digit) they are plotted in orange, and in blue otherwise. (diagonal entries) Histogram of distances calculated from each embedding. }
  \label{fig:mnist20_centroids}
\end{sidewaysfigure}

\begin{sidewaysfigure}
  \centering
  \includegraphics[width=\textwidth]{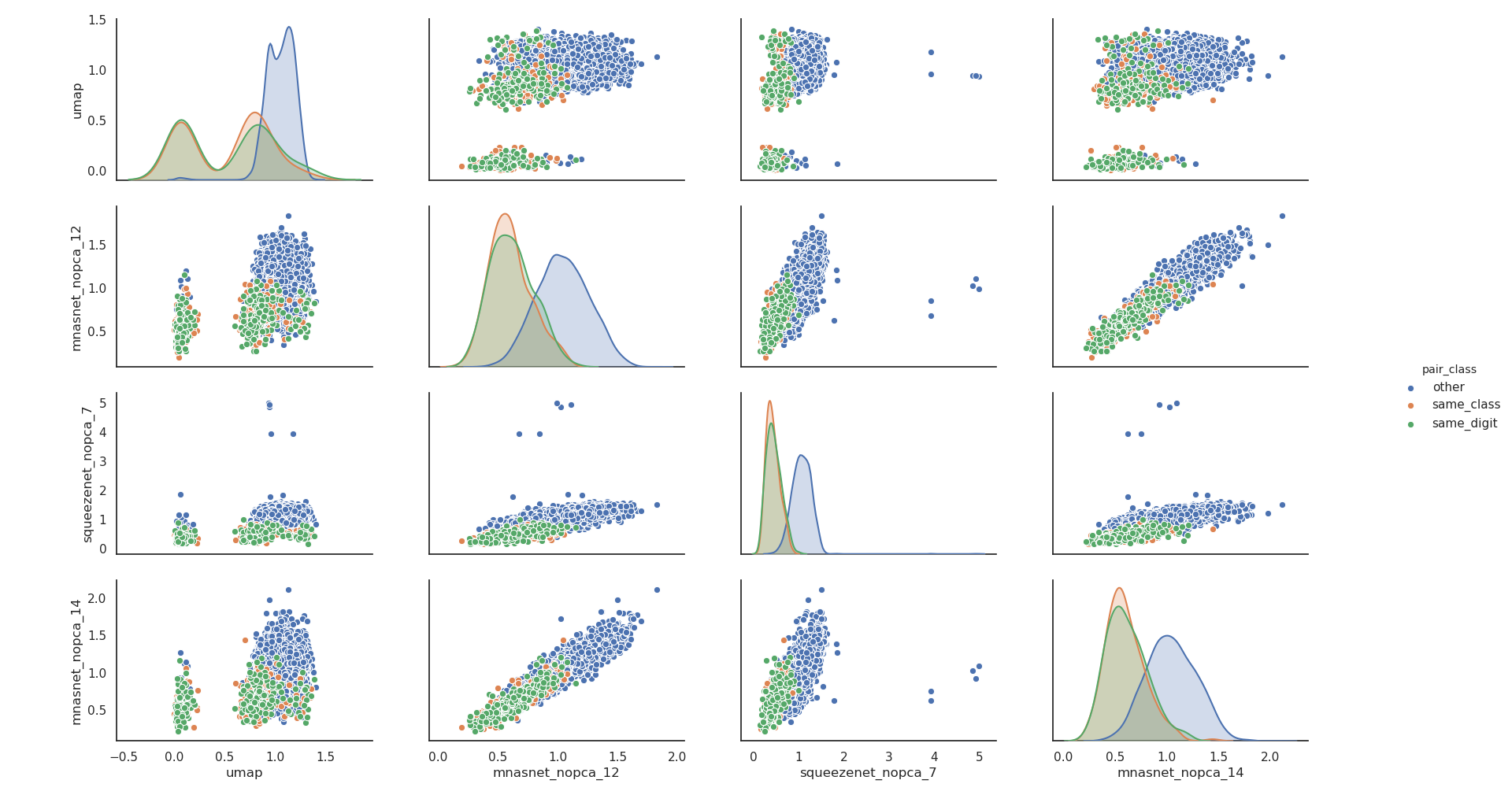}
  \caption{Distances between test images in MNIST-20. (off-diagonal) Each point represents distances between two images, calculated from different embeddings in x and y axis. If the images are in ``same\_class'' (the images have the same label) they are plotted in orange. If they are in ``same\_digit'' (they have different labels but they correspond to the same digit) they are plotted in green. Otherwise, they are plotted in blue. (diagonal entries) Histogram of distances calculated from each embedding.}
  \label{fig:mnist20_test}
\end{sidewaysfigure}

\begin{table}
\caption{Results MNIST-20 dataset}
  \centering
  \begin{tabular}{lllll}
    \toprule
Selection &    Model     & Accuracy     & Confusion same digits (centroids) & Confusion same digits (test data)\\
\midrule
& LDA on pixels  & 0.433  &  \textbf{0.000} & 0.490 \\
\midrule
Corrected  & supervised UMAP      &  -     &  0.178 &  \bf{0.239} \\
centroid   &  CNN-LDA (best)             &  0.473 &  \bf{0.000} &  0.299 \\
shift      &   CNN-LDA (mean best 5)     &  0.477 &  \bf{0.000} &  \bf{0.241} \\
& PCA + LDA & 0.438 & \textbf{0.000} & 0.500 \\
\midrule
Centroid  & supervised UMAP       &  -     &  0.222 & \bf{0.266} \\
shift   &  CNN-LDA (best)         &  0.462 &  \bf{0.000} & 0.306 \\
      &  CNN-LDA (mean best 5)   &  0.465 &  \bf{0.001} & 0.310 \\
& PCA + LDA & 0.438 & \textbf{0.000} & 0.500 \\
\bottomrule
  \end{tabular}
  \label{tab:mnist20}
\end{table}

We compared our results against supervised UMAP, as implemented in \textit{github.com/lmcinnes/umap} \cite{mcinnes2018umap-software}. The LDA applied to the CNN features finds that the centroids of classes corresponding to the same digit (``same\_digit'', e.g. 2 and 12) lay much closer than classes corresponding to different digits (``other'', e.g. 1 and 14) (\FIG{mnist20best}). In most of our models, the minimum distance between centroids of different digits  (``other'') is larger than the maximum distance between centroids representing same digits  (``same\_digit''), producing a confusion score of $0.0$ (\FIG{mnist20_centroids}, \TABLE{mnist20}). In comparison, UMAP produces distances between centroids different digits that are sometimes smaller than distances between centroids of same digits, leading to positive confusion scores (\FIG{mnist20_centroids}, \TABLE{mnist20}).

Distances between points in the test dataset are divided in three classes: ``same\_class'' if they are from the same class (assigned the same label), ``same\_digit'' if they are from different classes but represent the same digit, and ``other''. In \FIG{mnist20_test} the different classes correspond to the colours orange, green and blue. Distributions of ``same\_class'' and ``same\_digit'' are almost the same, as expected because label duplication is performed at random. Distances in ``other'' are larger than in the other two classes, but there is an overlap (\FIG{mnist20_test}, diagonal entries). All methods produce a similar overlap between ``same\_digit'' and ``other'', UMAP being the best. 

Please note how CNN-LDA produces a unimodal distribution, while UMAP produces a bimodal distribution in classes ``same class'' and ``same digit''. This suggests that UMAP separates each class in a cluster, while CNN-LDA produces a cluster for each digit.

\subsection{Butterfly dataset}

\begin{table}
\caption{Results butterfly dataset. Model selection is the corrected centroid shift}
  \centering
  \begin{tabular}{lllll}
    \toprule
Selection &    Model     & Accuracy     & Confusion comimics (centroids) & Confusion comimics (test data)\\
\midrule
Corrected  & supervised umap             &  -         &  0.632      &  0.957 \\
centroid   &  CNN-LDA (best)             & \bf{0.888} &  0.325      &  0.503 \\
shift      & CNN-LDA (mean best 5)       & 0.887      &  0.350      &  0.509 \\
           & CNN-PCA-LDA (best)          & 0.877      &  \bf{0.269} & 0.452 \\
           & CNN-PCA-LDA (mean best 5)   & 0.886      &  0.340      & 0.489 \\
           & PCA + LDA (best)            & 0.792      & 0.275       &  0.574 \\
           & PCA + LDA (mean best 5)     & 0.765      & \bf{0.251}  &  0.621 \\
\midrule
Centroid  & supervised umap          &  -             & 0.667     &        0.760 \\
shift   &   CNN-LDA (best)           &  0.858         & 0.301     & 0.509 \\
        & CNN-LDA (mean best 5)      & \textbf{0.860} &  0.312    &        0.506 \\
        &  CNN-PCA-LDA (best)        &  0.763         &  \textbf{0.219} &  0.505 \\
        &  CNN-PCA-LDA (mean best 5) &  0.759         &  \textbf{0.229}  & 0.516 \\
        & PCA + LDA                  & 0.731          & 0.254     &      0.633 \\
        & PCA + LDA (mean best 5)     & 0.765      & 0.251  &  0.621 \\
\midrule
  & 	 butterflynet &  0.860 &  0.354 &  \bf{0.314} \\

\bottomrule
  \end{tabular}
  \label{tab:butterfly}
\end{table}

\begin{sidewaysfigure}
  \centering
  \includegraphics[width=\textwidth]{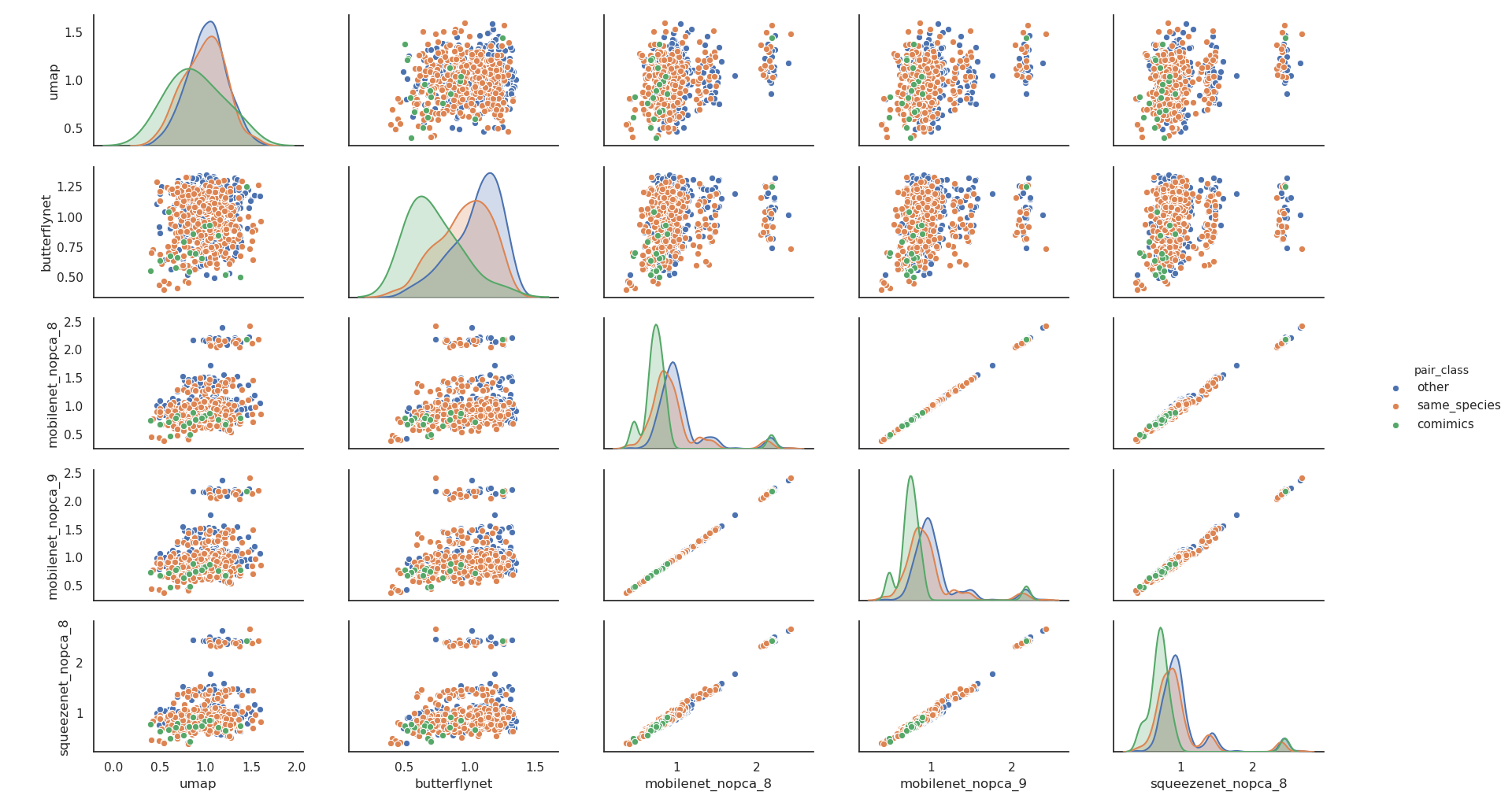}
  \caption{Distances between centroids in the butterfly dataset. (off-diagonal) Each point represents distances between two centroids, calculated from different embeddings in x and y axis. If the pair of centroids correspond to the class ``comimics'' (subspecies in a comimics complex) they are plotted in green, if the pair of centroids correspond to the class ``same\_species'' (subespecies in the same species, but not in the same comimics complex) they are plotted in orange, and in blue otherwise (different species and not in the same comimic complex). (diagonal entries) Histogram of distances calculated from each embedding. }
  \label{fig:dist_centroids}
\end{sidewaysfigure}

\begin{sidewaysfigure}
  \centering
  \includegraphics[width=\textwidth]{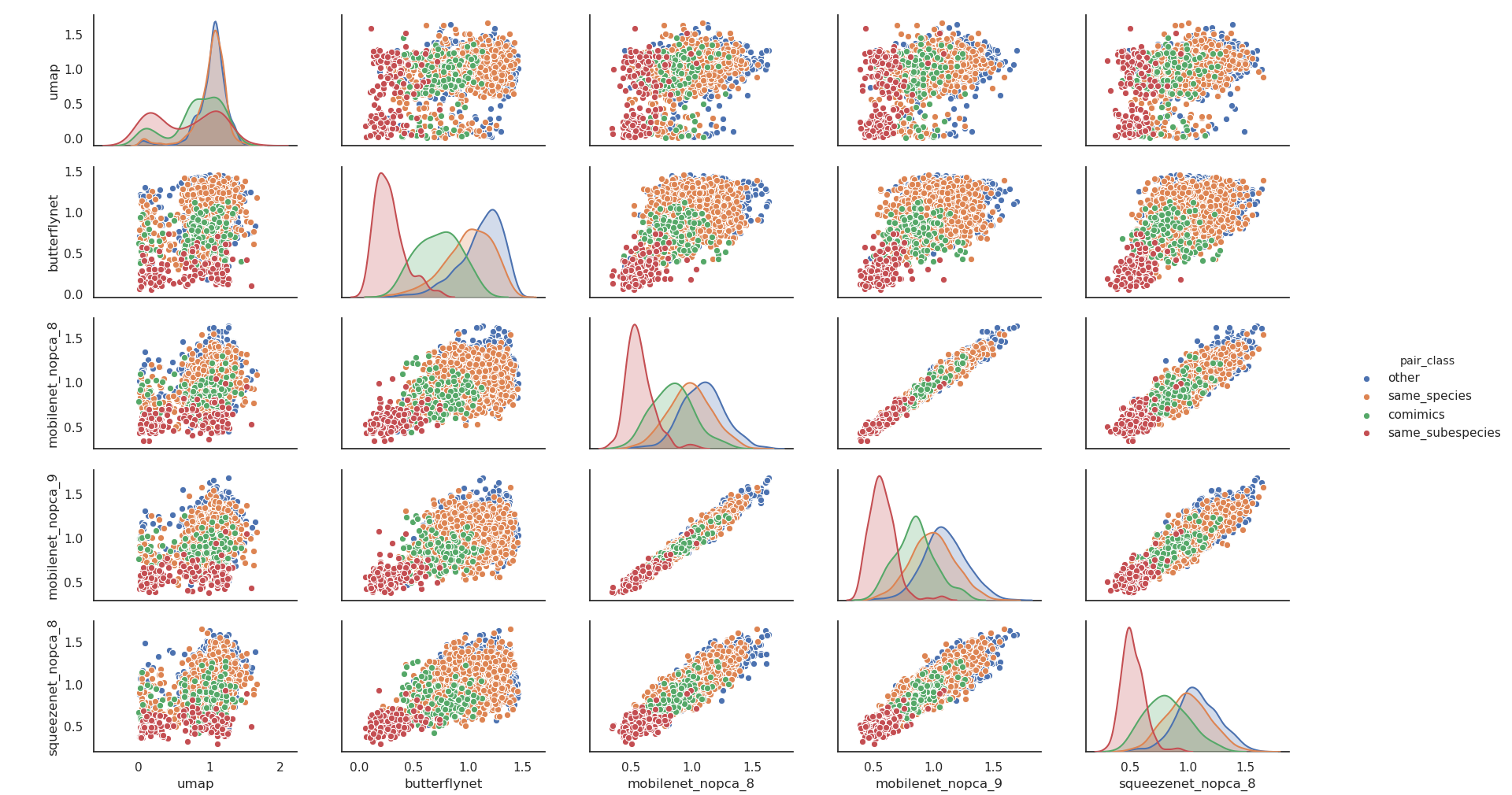}
  \caption{Distances between images in the butterfly dataset. (off-diagonal) Each point represents distances between two images, calculated from different embeddings in x and y axis. If the pair of images are from the same subspecies they are plotted in red. If the pair of images correspond to the class ``comimics'' (images in different subspecies of the same comimics complex) they are plotted in green, if the pair of images correspond to the class ``same\_species'' (images in different subespecies, in the same species, but not in the same comimics complex) they are plotted in orange, and in blue otherwise (images from different subspecies, different species and not in the same comimic complex). (diagonal entries) Histogram of distances calculated from each embedding.}
  \label{fig:dist_test}
\end{sidewaysfigure}

We tested our method (LDA-CNN) on a high-quality photographic dataset of butterfly photographs, obtained and curated by \cite{cuthill2019deep}. The interest of this dataset is that there is a hierarchy of labels for each butterfly specimen. The highest hierarchy in labelling is species, of which we have two, (\textit{H. erato} and \textit{H. melpomene}). Each species is divided in several subspecies, producing a total of 38 subspecies. Butterflies in the same species do not necessarily look similar but they do if they belong to the same subspecies. In addition, biologists agree that some of the subspecies look similar to other subspecies (they are in the same comimic complex), and they have been labelled to be ``comimics''. 

It is this hierarchy, where some classes are considered to be more similar than others, that allows us to test our approach. We produce an embedding of the butterfly images where specimens of the same subspecies fall in the same place. To test the quality of the embedding, we assess whether comimics are closer than other pairs of specimens.

The dataset is composed of two pictures (dorsal and ventral) for each of the 1234 specimens \cite{cuthill2019deep}. All the photographs were taken with consistent conditions, and resized to a height of 64 pixels (keeping the original aspect ratio). We have followed the same train-test split as in \cite{cuthill2019deep} (1500/968), and we performed the same data augmentation on our training set (combinations of 0, 1 and 2 horizontal and vertical pixel shifts).

We compared our results to the approach followed in \cite{cuthill2019deep}. In this approach, a 64-dimensional embedding was built by training a network, ButterflyNet, minimising a triplet loss. The result is that images from the same subspecies where pulled together, while images from different subspecies are pushed away. We used the result of a single training run as made public with the article \cite{cuthill2019deep}. As above, we also compared our results against supervised UMAP.

We performed LDA on the CNN features. We calculated the average position of each class in the training set in \cite{cuthill2019deep} and the centroid of each class in our embedding (parameters obtained during the LDA training). We used the test set to select the best networks. We selected the networks that produced the smallest ``centroid shift'' (distance between class centroids in train and test datasets, see Methods). The accuracies of the best networks are similar to the accuracy reported in \cite{cuthill2019deep}. 

We calculated the pairwise distances between all subespecies' centroids. We divided them in three groups: 'comimic' if the subespecies are in the same comimic complex, 'same-species' if they are not comimic but they belong to the same species (\textit{H. erato} or \textit{H. melpomene}), or 'other' if they are not comimics and each one belongs to a different species. We found the separation obtained between comimic subspecies and 'other' to be similar or slightly larger to that found by \cite{cuthill2019deep} (see \TABLE{butterfly}, \FIG{dist_centroids}) We do not seem to obtain more phylogenetic signal than \cite{cuthill2019deep} (the orange and blue curves of the histograms in \FIG{dist_centroids} have a similar overlap). Note the high correlation between different distances given by CNN-LDA produced by different CNN or different layers (\FIG{dist_centroids}). This high correlation persists in the other high-performing CNN-LDA we tested (other combinations of CNN and layer). Supervised UMAP performs the worst, placing comimics as far away as non comimics.

Instead of using the plain centroid shift to select the best model, we can first scale each dimension in the embedding by the standard deviation of centroids location along that dimension. By doing that, we punish shifts along dimensions that produce classification change. We found that the models thus selected have higher classification accuracy, but their ability to separate comimics decreases slightly (see \TABLE{butterfly}).

We calculated the pairwise distances between all images. We divided them in four groups. The first group is 'same\_subespecies' if the images correspond to the same subspecies. Otherwise, the pair of images is classified as above: 'comimics' if the subespecies are in the same comimic complex, 'same\_species' if they are not comimic but they belong to the same species, or 'other' otherwise. We found the separation obtained between comimic subspecies and subespecies of the same species to be slightly worse to that found by \cite{cuthill2019deep} (see \TABLE{butterfly}, \FIG{dist_test}). Supervised UMAP performs the worst, sometimes placing images of the same subspecies far away.

\section{Discussion}

We found that, for the two datasets studied, performing linear discriminant analysis (LDA) on pretrained convoluntional features enables supervised dimensionality reduction with a similar quality to existing methods. 

LDA has been used before as a classifier on top of a pre-trained convolutional neural networks (CNN) \cite{el2016face}. In addition, there are methods that allow training a CNN and an LDA together in an end-to-end fashion to produce a classifier \cite{dorfer2015deep}. However, to our knowledge none of the previous articles apply this architecture to a supervised dimensionality reduction problem.

Contrastive losses (e.g. siamese and triplet networks) \cite{hadsell2006dimensionality, schroff2015facenet, cuthill2019deep} are popular approaches to supervised dimensionality reduction. They are designed to iteratively pull together the embeddings of elements of the same class, while pushing away the embeddings of elements of different classes. After training, elements of the same class are together in the space, and elements of different classes are not. In addition, the resulting embedding usually succeeds in placing elements of similar classes in close points of the embedding. It is important to note that this is not produced by any imposed condition that we can explicitly control. It is ultimately due to inductive biases of neural networks, favouring smoother embeddings that do not separate images that are close in pixel space. We suggest that a research line worth pursuing is the use of explicit constraints on the smoothness of the embedding, for example placing an upper bound on the Lipschitz constant \cite{gouk2018regularisation}.

The limitations of our study can be used to extend our results in several directions. First, we have only made a comparison with a subset of state-of-the-art methods, but a larger variety of alternative methods exist. For instance, we compared against the results of \cite{cuthill2019deep}, but not against other approaches using contrastive losses \cite{hadsell2006dimensionality, schroff2015facenet}. Our tests were performed in two datasets, one created to have ground data and the other in a dataset of buttefly images relevant to biologists. While results are encouraging for these two datasets, the usefulness of the approach taken needs to be tested in other problems. The current implementation is restricted to relatively few data points. This is because the algorithms we use to perform PCA and LDA use all data points at the same time, quickly running out of memory if the images or the datasets are large. However, our method can be modified for use in larger datasets by using algorithms that iteratively calculate PCA and LDA from minibatches (e.g. \cite{li2004incremental, pang2005incremental}). The advantage of using LDA is that is a transformation that cannot deform the space, and it is restricted to a linear transformation (sphering) and a projection to a subspace. It also provides reasonable and meaningful units in each of the dimensions (intra-class standard deviation). However, note that the CNN makes very complex transformations and we are assuming that are meaningful for our analysis. A formal justification of this assumption is still an open problem.

Despite these limitations, we believe our results show that LDA applied to a CNN can be a principled approach to supervised dimensionality reduction with results comparable to those of other approaches for which a principled understanding might be more difficult to obtain.

\section{Code availability}

The code has been released as free software. It can be found in \url{https://gitlab.com/polavieja_lab/cnn-lda}

\bibliographystyle{unsrt}  
\bibliography{references}  

\begin{thebibliography}{10}

\bibitem{schroff2015facenet}
Florian Schroff, Dmitry Kalenichenko, and James Philbin.
\newblock Facenet: A unified embedding for face recognition and clustering.
\newblock In {\em Proceedings of the IEEE conference on computer vision and
  pattern recognition}, pages 815--823, 2015.

\bibitem{hastie2009elements}
Trevor Hastie, Robert Tibshirani, and Jerome Friedman.
\newblock {\em The elements of statistical learning: data mining, inference,
  and prediction}.
\newblock Springer Science \& Business Media, 2009.

\bibitem{sharif2014cnn}
Ali Sharif~Razavian, Hossein Azizpour, Josephine Sullivan, and Stefan Carlsson.
\newblock Cnn features off-the-shelf: an astounding baseline for recognition.
\newblock In {\em Proceedings of the IEEE conference on computer vision and
  pattern recognition workshops}, pages 806--813, 2014.

\bibitem{el2016face}
Hachim El~Khiyari and Harry Wechsler.
\newblock Face recognition across time lapse using convolutional neural
  networks.
\newblock {\em Journal of Information Security}, 7(3):141--151, 2016.

\bibitem{lopes2017pre}
UK~Lopes and Jo{\~a}o~Francisco Valiati.
\newblock Pre-trained convolutional neural networks as feature extractors for
  tuberculosis detection.
\newblock {\em Computers in biology and medicine}, 89:135--143, 2017.

\bibitem{verma2014using}
Ankit Verma and Lovekesh Vig.
\newblock Using convolutional neural networks to discover cogntively validated
  features for gender classification.
\newblock In {\em 2014 International Conference on Soft Computing and Machine
  Intelligence}, pages 33--37. IEEE, 2014.

\bibitem{lindsay2020convolutional}
Grace~W Lindsay.
\newblock Convolutional neural networks as a model of the visual system: Past,
  present, and future.
\newblock {\em arXiv preprint arXiv:2001.07092}, 2020.

\bibitem{scikit-learn}
F.~Pedregosa, G.~Varoquaux, A.~Gramfort, V.~Michel, B.~Thirion, O.~Grisel,
  M.~Blondel, P.~Prettenhofer, R.~Weiss, V.~Dubourg, J.~Vanderplas, A.~Passos,
  D.~Cournapeau, M.~Brucher, M.~Perrot, and E.~Duchesnay.
\newblock Scikit-learn: Machine learning in {P}ython.
\newblock {\em Journal of Machine Learning Research}, 12:2825--2830, 2011.

\bibitem{bianco2018benchmark}
Simone Bianco, Remi Cadene, Luigi Celona, and Paolo Napoletano.
\newblock Benchmark analysis of representative deep neural network
  architectures.
\newblock {\em IEEE Access}, 6:64270--64277, 2018.

\bibitem{iandola2014densenet}
Forrest Iandola, Matt Moskewicz, Sergey Karayev, Ross Girshick, Trevor Darrell,
  and Kurt Keutzer.
\newblock Densenet: Implementing efficient convnet descriptor pyramids.
\newblock {\em arXiv preprint arXiv:1404.1869}, 2014.

\bibitem{szegedy2015going}
Christian Szegedy, Wei Liu, Yangqing Jia, Pierre Sermanet, Scott Reed, Dragomir
  Anguelov, Dumitru Erhan, Vincent Vanhoucke, and Andrew Rabinovich.
\newblock Going deeper with convolutions.
\newblock In {\em Proceedings of the IEEE conference on computer vision and
  pattern recognition}, pages 1--9, 2015.

\bibitem{tan2019mnasnet}
Mingxing Tan, Bo~Chen, Ruoming Pang, Vijay Vasudevan, Mark Sandler, Andrew
  Howard, and Quoc~V Le.
\newblock Mnasnet: Platform-aware neural architecture search for mobile.
\newblock In {\em Proceedings of the IEEE Conference on Computer Vision and
  Pattern Recognition}, pages 2820--2828, 2019.

\bibitem{sandler2018mobilenetv2}
Mark Sandler, Andrew Howard, Menglong Zhu, Andrey Zhmoginov, and Liang-Chieh
  Chen.
\newblock Mobilenetv2: Inverted residuals and linear bottlenecks.
\newblock In {\em Proceedings of the IEEE conference on computer vision and
  pattern recognition}, pages 4510--4520, 2018.

\bibitem{ma2018shufflenet}
Ningning Ma, Xiangyu Zhang, Hai-Tao Zheng, and Jian Sun.
\newblock Shufflenet v2: Practical guidelines for efficient cnn architecture
  design.
\newblock In {\em Proceedings of the European Conference on Computer Vision
  (ECCV)}, pages 116--131, 2018.

\bibitem{iandola2016squeezenet}
Forrest~N Iandola, Song Han, Matthew~W Moskewicz, Khalid Ashraf, William~J
  Dally, and Kurt Keutzer.
\newblock Squeezenet: Alexnet-level accuracy with 50x fewer parameters and< 0.5
  mb model size.
\newblock {\em arXiv preprint arXiv:1602.07360}, 2016.

\bibitem{pasze2019pytorch}
Adam Paszke, Sam Gross, Francisco Massa, Adam Lerer, James Bradbury, Gregory
  Chanan, Trevor Killeen, Zeming Lin, Natalia Gimelshein, Luca Antiga, Alban
  Desmaison, Andreas Kopf, Edward Yang, Zachary DeVito, Martin Raison, Alykhan
  Tejani, Sasank Chilamkurthy, Benoit Steiner, Lu~Fang, Junjie Bai, and Soumith
  Chintala.
\newblock Pytorch: An imperative style, high-performance deep learning library.
\newblock In {\em Advances in Neural Information Processing Systems 32}. Curran
  Associates, Inc., 2019.

\bibitem{mcinnes2018umap-software}
Leland McInnes, John Healy, Nathaniel Saul, and Lukas Grossberger.
\newblock Umap: Uniform manifold approximation and projection.
\newblock {\em The Journal of Open Source Software}, 3(29):861, 2018.

\bibitem{cuthill2019deep}
Jennifer F~Hoyal Cuthill, Nicholas Guttenberg, Sophie Ledger, Robyn Crowther,
  and Blanca Huertas.
\newblock Deep learning on butterfly phenotypes tests evolution’s oldest
  mathematical model.
\newblock {\em Science advances}, 5(8):eaaw4967, 2019.

\bibitem{dorfer2015deep}
Matthias Dorfer, Rainer Kelz, and Gerhard Widmer.
\newblock Deep linear discriminant analysis.
\newblock {\em arXiv preprint arXiv:1511.04707}, 2015.

\bibitem{hadsell2006dimensionality}
Raia Hadsell, Sumit Chopra, and Yann LeCun.
\newblock Dimensionality reduction by learning an invariant mapping.
\newblock In {\em 2006 IEEE Computer Society Conference on Computer Vision and
  Pattern Recognition (CVPR'06)}, volume~2, pages 1735--1742. IEEE, 2006.

\bibitem{gouk2018regularisation}
Henry Gouk, Eibe Frank, Bernhard Pfahringer, and Michael~J Cree.
\newblock Regularisation of neural networks by enforcing lipschitz continuity.
\newblock {\em stat}, 1050:14, 2018.

\bibitem{li2004incremental}
Yongmin Li.
\newblock On incremental and robust subspace learning.
\newblock {\em Pattern recognition}, 37(7):1509--1518, 2004.

\bibitem{pang2005incremental}
Shaoning Pang, Seiichi Ozawa, and Nikola Kasabov.
\newblock Incremental linear discriminant analysis for classification of data
  streams.
\newblock {\em IEEE transactions on Systems, Man, and Cybernetics, part B
  (Cybernetics)}, 35(5):905--914, 2005.

\end{thebibliography}

\end{document}